\documentclass{article}

\usepackage{arxiv}

\usepackage{lmodern}

\usepackage[utf8]{inputenc}
\usepackage[T1]{fontenc}
\usepackage{hyperref}
\usepackage{url}
\usepackage{booktabs}
\usepackage{array}
\usepackage{amsmath}
\usepackage{microtype}
\usepackage{graphicx}
\usepackage{flafter}
\usepackage{placeins}
\usepackage{subcaption}

\usepackage[numbers,sort&compress]{natbib}
\usepackage{xcolor}
\usepackage{orcidlink}

\renewcommand{\undertitle}{}
\date{}

\title{Lambda-Hold Control: Human-Like Movement Emerges from a Minimal
Task Reward in Predictive Musculoskeletal Simulation}

\author{
  {\large Jun Hyuk Lee\,\orcidlink{0000-0003-3345-5303}\,$^{1}$ \hspace{3em}
  Chihyeong Lee\,\orcidlink{0009-0002-4335-9825}\,$^{1}$ \hspace{3em}
  Jooeun Ahn\,\orcidlink{0000-0002-7964-5148}\,$^{1,2,3,*}$} \\[2pt]
  {\normalfont\small $^{1}$Department of Physical Education, Seoul National University, Seoul, Republic of Korea}\\
  {\normalfont\small $^{2}$Institute of Sport Science, Seoul National University, Seoul, Republic of Korea}\\
  {\normalfont\small $^{3}$SNU Robotics Institute, Seoul National University, Seoul, Republic of Korea}\\
  {\normalfont\small $^{*}$Corresponding author: \texttt{ahnjooeun@snu.ac.kr}}\\[4pt]
  {\normalfont\small Project page and video demos: \url{https://lee-jun-hyuk-37.github.io/projects/lambda-hold/}}
}

\begin{document}
\maketitle
\vspace{-8mm}

\begin{abstract}
The massive overactuation in the human musculoskeletal system makes it challenging to train musculoskeletal models to generate human-like motion via reinforcement learning, primarily because exploration in the resulting high-dimensional and redundant action space is extremely inefficient.
To address this problem, we propose the $\lambda$-hold controller, inspired by the equilibrium-point (EP) hypothesis, which has been widely supported by extensive evidence from human motor control studies.
The policy’s control variable is the per-muscle EP threshold length $\lambda$, from which a stretch-reflex recruitment law computes the muscle excitations automatically.
Holding each $\lambda$ over an interval of the gait phase also sharply reduces the frequency at which the policy must be queried.
Consequently, the controller, to our knowledge for the first time, enables a muscle-actuated skeletal model to learn human-like sprinting using only a minimal reward within an hour of training.
The efficient exploration through the proposed $\lambda$-hold controller is not merely an engineering trick but an approach grounded in physiology, bringing together the EP hypothesis, intermittent control, and optimal feedback control.
Beyond encapsulating human-like behavior in predictive simulation, this achievement contributes to developing a learnable model of the human motor controller.
\end{abstract}


\section{Introduction}
\label{sec:intro}

\subsection{Predictive simulation of human motor control}

Simulations of human movement fall into two broad families that differ in the direction of inference.
Inverse simulation starts from experimentally measured motion and estimates the internal quantities, such as joint moments, muscle forces, and activations, that could have produced it.
Predictive simulation reverses this direction.
It is solved through forward dynamics but takes no experimental trajectory as input, generating human-like behavior from a model of the body and its controller and using experimental data only afterwards, for validation \citep{falisse2019,geijtenbeek2019}.
Predictive simulation is an act of modeling rather than curve fitting, and the two differ in two respects.
The first is generalization.
A fitted curve reproduces the data it was tuned on, whereas a model that captures the generative mechanism extrapolates to conditions it never saw.
The second is interpretability.
A fitted curve's parameters carry no physiological referent, whereas a model's internal variables and hypotheses correspond to physiological quantities, and can be examined directly; the model can suggest feasible explanation of the behavior rather than merely reproducing it.

Predictive simulations of human motor control differ in the level of abstraction at which the controller operates.
At the joint-torque level the controller commands net joint moments directly, and most work remains here, especially in humanoid robotics and physics-based character animation \citep{heess2017,peng2018deepmimic,peng2021amp,yu2018symmetry}.
Below it, the muscle-activation level acts on individual muscle--tendon units.
Below that, the neural-circuit level acts on spinal and supraspinal pathways such as reflex modules and central pattern generators (CPGs).
Deeper still, the cellular level resolves the biophysical dynamics of neurons and synapses.
This work sits between the muscle and neural-circuit levels: it resolves a long-standing difficulty at the muscle level through a simple piece of neural-circuit-level modeling.

\subsection{The muscle-redundancy problem}

Control at the muscle level is not merely more difficult than at the joint-torque level.
It is a different kind of problem.
The human lower limb is actuated by far more muscles than the limb's mechanical degrees of freedom, creating a redundancy with two consequences.
The first consequence is the curse of dimensionality.
The second is that the mapping from activation to movement is many-to-one, since the same kinematics and kinetics can arise from entirely different activation profiles.
Recovering the activations underlying a given motion is therefore ill-posed \citep{erdemir2007}.

Optimization-based approaches resolve this redundancy by imposing an optimality criterion.
The classic choice is minimization of effort, typically quantified as the sum of squared or cubed activations \citep{crowninshield1981,thelen2003cmc}.
A more physiological alternative is to minimize a proxy for metabolic cost \citep{umberger2010,uchida2016}.
Such criteria yield a unique solution, but cannot guarantee that it is the one the nervous system uses, and the resulting activations often correlate only weakly with measured electromyography (EMG) \citep{shuman2019}.
This gap between a mathematically convenient optimum and the biological observation is the clear limitation of the control at the muscle level to date.

\subsection{The inefficient-exploration problem}

Viewed from the reinforcement-learning (RL) side, this redundancy reappears as what we call the inefficient-exploration problem.
The inefficiency has two sources.
The first is the structure of the excitation-to-joint map.
An RL agent explores by perturbing each muscle's excitation independently at every timestep, as standard Gaussian action noise does.
Because the map is many-to-one, these uncorrelated perturbations largely cancel in joint space and produce little variation in behavior \citep{schumacher2023deprl,chiappa2023}.
Synergist muscles that should act together are perturbed incoherently, so exploration in excitation space rarely becomes exploration of joint space.
The second source is the high dimensionality of the action space, requiring the policy to specify a full muscle excitation vector hundreds of times per second.
Fundamentally, humans do not control every muscle independently at every instant \citep{davella2003}, yet a naive RL agent is forced to.

\subsection{Prior approaches}
\label{sec:prior}

Four families of methods make muscle-level exploration tractable, each addressing the redundancy differently.

\textbf{Imposing a fixed rule and tuning its gains.}
Reflex-based controllers prescribe a hand-designed feedback structure and optimize only its gains \citep{geyer2010,song2015}.
The predefined structure makes the search tractable while still allowing a wide range of locomotion behaviors to be composed from spinal reflex modules.
The resulting control, however, remains bound by a structure specified in advance rather than learned.

\textbf{Shaping the exploration noise.}
The excitation action space is left unchanged, and the uncorrelated Gaussian noise is replaced with temporally or spatially correlated noise such as Ornstein--Uhlenbeck noise \citep{lillicrap2016ddpg}, pink noise \citep{eberhard2023pink}, or state-dependent exploration \citep{ruckstiess2008sde,raffin2021gsde}.
The correlation reduces independent exploration across muscles or across time, which makes the excitation space more tractable without reducing its dimensionality.
However, the correlation is injected externally at the exploration stage rather than arising from the controller's own commands, and the gain over plain Gaussian noise is not guaranteed \citep{fujimoto2018td3}.

\textbf{Injecting a self-organizing exploration signal.}
The noise is replaced with a self-organizing drive, as in DEP-RL \citep{schumacher2023deprl}, which has since been used to learn natural, robust muscle-driven walking from biologically plausible objectives without motion-capture demonstrations \citep{schumacher2025emergence}.
Its differential-plasticity signal exploits an effect closely related to the stretch reflex, exciting muscles that have lengthened and relaxing those that have shortened, so the perturbations it generates are structured by the mechanical state rather than independent across muscles.
This reflex-like coupling, however, acts only during data collection.
It is a separate controller that intermittently takes over to fill the replay buffer, and the trained policy still outputs raw excitations.

\textbf{Reducing the dimension using muscle synergies.}
The action is projected onto a low-dimensional synergy basis, and the policy commands only the synergy coefficients \citep{park2026}.
The basis is either extracted from measured muscle activity \citep{chvatal2013} or distilled from a trained policy \citep{berg2023sar}.
Replacing individual muscle commands with a smaller set of coefficients reduces the dimensionality of the action space substantially, but the synergy matrix is fixed, so the mapping from latent commands to muscle excitations cannot adapt online, unlike the task- and state-dependent synergies observed in humans.

In all four approaches, coordination is imposed externally, whether designed into the circuitry, injected into the exploration noise, produced by a separate controller during data collection, or fixed in a predefined synergy basis.

\subsection{Contributions}

We resolve the inefficient-exploration problem by changing the control variable.
Instead of commanding muscle excitations, the policy commands per-muscle threshold lengths $\lambda$, the control variable of Feldman's equilibrium-point (EP) hypothesis \citep{feldman1986,gribble1998}.
A stretch-reflex recruitment law converts each $\lambda$, together with the muscle's evolving length and velocity, into an excitation.
Each muscle responds only to its own length, but the skeleton couples those lengths mechanically, so independent perturbations in $\lambda$-space produce excitations that covary across muscles.
Coordination is therefore not imposed externally but emerges from the body itself.

Günther and Ruder showed that hand-tuned $\lambda$ can drive predictive gait \citep{gunther2003}.
We are, to our knowledge, the first to use deep RL to select $\lambda$ in a high-fidelity predictive musculoskeletal simulation, and the first to hold each $\lambda$ over a gait-phase interval, so that the policy is queried only at sparse, phase-locked decision points while the stretch reflex fills the gaps between them.
The reflex is therefore part of the control law rather than an addition to training.
In DEP-RL the reflex-like rule takes over from the policy during data collection, whereas in $\lambda$-hold, the reflex-like rule is what the policy commands throughout both training and deployment.

Our contributions are:

\begin{itemize}
  \item \textbf{Human-like sprinting from a minimal reward.}
        We produced, to our knowledge, the first coordinated, human-like sprinting in a predictive musculoskeletal simulation using a reward consisting of a forward-velocity term alone.
        Prior muscle-level controllers rely on reference-motion imitation \citep{lee2019scalable,peng2018deepmimic}, on effort or metabolic-cost minimization combined with posture, joint-limit, and stability terms \citep{song2021,falisse2019,ong2019,caggiano2022myosuite,park2026,schumacher2025emergence}, on hand-designed, locomotion-specific reflex circuitry with tuned gains \citep{geyer2010,song2015}, on training curricula \citep{caggiano2022myosuite,heess2017,park2026}, or on gait-symmetry aids such as policy mirroring \citep{park2026} or an explicit symmetry reward \citep{yu2018symmetry}.
        Our controller uses none of these.
  \item \textbf{Resolving the inefficient-exploration problem.}
        Because $\lambda$-hold and excitation-level control are queried at different rates, we compare them against simulation steps rather than decision steps.
        On that axis, $\lambda$-hold covers the joint-space state distribution earlier and more broadly, and needs far fewer policy decisions to do so.
        The substantial gain in learning speed enables a sprint to emerge within about an hour of wall-clock training on a single workstation (Section~\ref{sec:res-learning}), so batches of hypotheses that were previously out of reach can be tested in a day.
  \item \textbf{Not an artificial but a physiologically grounded mechanism.}
        The recruitment law follows the stretch reflex as formalized in Feldman's EP hypothesis \citep{feldman1986,gribble1998}, a rule broadly supported by experimental evidence in human motor control, not a simulation artifact or an external exploration trick.
        The resulting controller is interpretable in neurophysiological terms.
\end{itemize}

Solving inefficient exploration matters on two fronts.
Scientifically, it moves predictive musculoskeletal simulation toward a principled account of how the human motor system selects among its redundant actuators.
Practically, applications from clinical planning and rehabilitation to physical human--robot interaction require a model that generalizes rather than merely fits, and one cheap enough to train to be adapted to an individual.
The $\lambda$-hold controller is one step toward a learnable model of the human motor controller.

\section{Methods}
\label{sec:methods}

\subsection{Musculoskeletal model and simulation}
\label{sec:model}

We use the H2190 three-dimensional musculoskeletal model, which has 21 mechanical degrees of freedom actuated by 90~muscle--tendon units, and simulate it in SCONE \citep{geijtenbeek2019} with the Hyfydy engine \citep{geijtenbeek2021}.
The index $i$ runs over the 90 muscles.
The simulation step is $0.01\,$s, and every method compared in this work shares it.
At each step the engine exposes each muscle's length $\ell_i$ and velocity $v_i$, and takes muscle excitation $e_i$ as its input.
In accordance with the default output format of the Hyfydy engine, both $\ell_i$ and $v_i$ are normalized by the optimal fiber length.

In this work we distinguish excitation from activation as follows.
The excitation $e_i \in [0,1]$ is the neural command the simulator receives, whereas the activation is the muscle's resulting active state, which the engine obtains from $e_i$ through the muscle's activation dynamics.
Muscle force then follows from the activation together with the muscle's force--length and force--velocity properties.

\subsection{The \texorpdfstring{$\lambda$}{λ}-hold controller}
\label{sec:lambda-hold}

\paragraph{Control variable: threshold length $\lambda$.}
The controller has two parts: a policy that outputs a per-muscle threshold length $\lambda_i$, and a stretch-reflex recruitment law that converts each $\lambda_i$ into an excitation $e_i$.
The threshold is the equilibrium point of the EP hypothesis \citep{feldman1986,gribble1998}, the muscle length at which the muscle begins to be recruited.
The law compares the current muscle length with the threshold and adds a rectified velocity term,
\begin{equation}
  e_i \;=\; \Big[\;
      G_{\mathrm{tonic}}\,\big(\ell_i - \lambda_i\big)
      \;+\; G_{\mathrm{phasic}}\,\big[\,v_i\,\big]_{+}
  \;\Big]_{0}^{1}
  \label{eq:reflex}
\end{equation}
where $[\,\cdot\,]_{0}^{1}$ clamps the result to $[0,1]$ and $[\,\cdot\,]_{+}=\max(0,\cdot)$.
The tonic and phasic terms respectively implement the static stretch reflex, a length feedback about the threshold, and the dynamic stretch reflex, a velocity feedback.
Rectifying the phasic term ensures that it adds drive only while the muscle is being stretched and never subtracts it during shortening \citep{proske2012}.

The reflex gains $G_{\mathrm{tonic}}$ and $G_{\mathrm{phasic}}$ are shared by all 90 muscles and fixed at 50 and 0.1 throughout this work; they may be tuned for other tasks.
The threshold $\lambda_i$ is the policy's command, bounded to $[0.6, 1.2]$.
A muscle below its threshold and not being stretched is silent, whereas one stretched beyond the threshold is recruited in proportion to the excess.

The law couples the muscles without containing any coupling term.
Each muscle responds only to its own length and velocity, but the skeleton makes those lengths covary, so muscles that lengthen together are recruited together.
Exploration in $\lambda$-space is therefore coordinated by construction rather than by a predefined synergy basis (Section~\ref{sec:disc-coverage}).

\begin{figure}[tb]
  \centering
  \includegraphics[width=\textwidth]{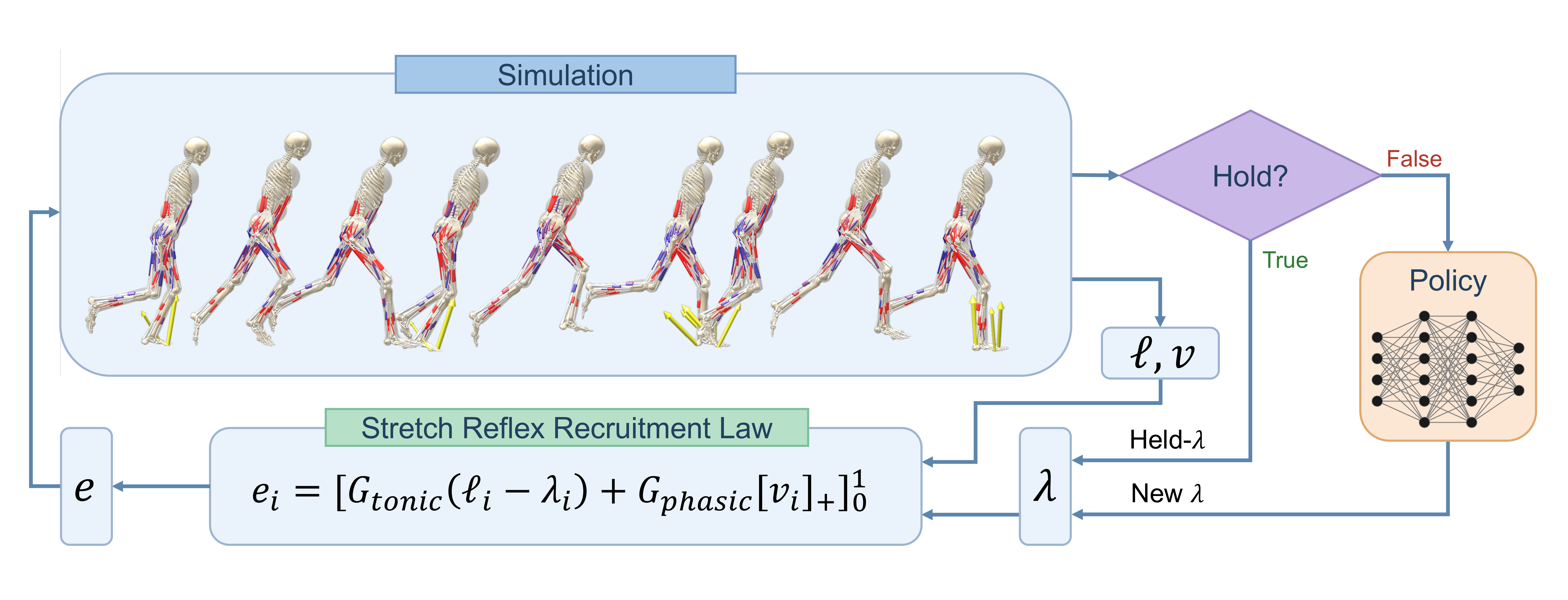}
  \caption{Overview of the $\lambda$-hold controller.
  The policy commands per-muscle threshold lengths $\lambda$ only at intermittent decision points.
  Between them $\lambda$ is held while a stretch-reflex recruitment law converts it, together with the evolving muscle length and velocity, into muscle excitation.
  Trained by reinforcement learning on a minimal reward of forward velocity alone, this controller learns human-like sprinting within about an hour of training.}
  \label{fig:lambda-overview}
\end{figure}

\paragraph{Holding $\lambda$ between decisions.}
The policy does not resample $\lambda$ at every simulation step.
It emits a new $\lambda$ at selected decision points and holds that value until the next one, while the recruitment law keeps generating excitation as $\ell_i$ and $v_i$ evolve.
We call the span between two decision points a hold interval.
It covers many simulation steps, and the reward is accumulated over it during training.
Because the decision points are locked to the gait phase (Section~\ref{sec:hold-timing}), hold intervals vary in duration.
The architecture therefore has two feedback rates. The policy sets a new target from the current state once per hold interval, and the reflex law tracks that target continuously in between (Figure~\ref{fig:lambda-overview}).

The policy is queried once per hold interval instead of once per simulation step.
As the hold interval increases, the number of policy decisions required during training decreases, reducing wall-clock training time.
Holding is not merely an engineering trick; we return to its physiological grounding in Section~\ref{sec:disc-holding}.
Nor is it equivalent to action repeat \citep{mnih2015,sharma2017} because the excitation keeps varying through the reflex loop throughout each interval.
This distinction is further clarified in Section~\ref{sec:baselines}.
How we place the decision points is treated next in Section~\ref{sec:hold-timing}.

\subsection{Decision timing}
\label{sec:hold-timing}

\begin{figure}[tb]
  \centering
  \includegraphics[width=\textwidth]{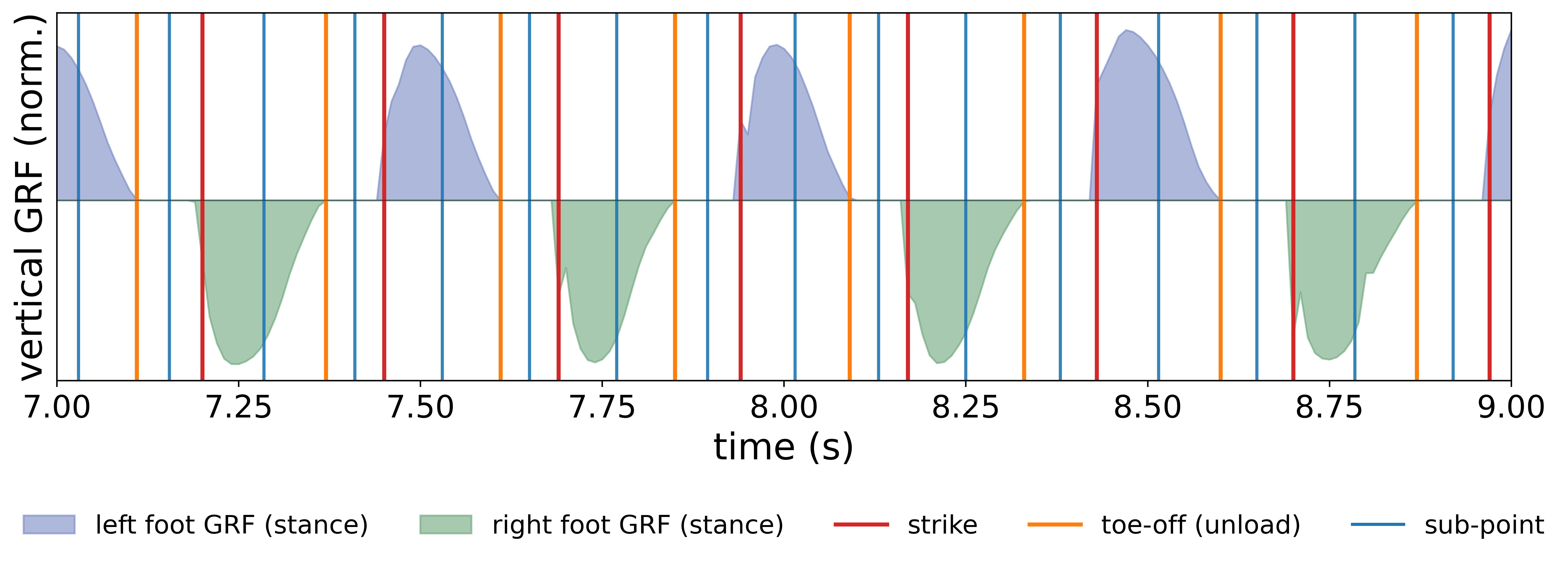}
  \caption{Decision timing during a high-speed $\lambda$-hold sprint. A new $\lambda$ is issued at each ground-reaction-force event, both foot strikes and toe-offs, with one additional decision point inserted between consecutive events. Filled curves are the per-foot vertical ground-reaction force (left foot up, right foot down). Vertical lines mark the $\lambda$ resample points: a foot strike (rising contact edge, red), a toe-off (falling edge, orange), and the interposed sub-point (blue). The policy is re-queried only at these instants. Between them, the held $\lambda$ and the stretch-reflex law generate the excitations.}
  \label{fig:timeline}
\end{figure}

We place the decision points at ground reaction force (GRF) events (Figure~\ref{fig:timeline}).
A new decision is triggered whenever either foot changes contact state, that is, whenever its vertical GRF crosses 0.05 body weight, whether rising (a foot strike) or falling (a toe-off).
Between two consecutive events we insert one additional decision point, the sub-point. We cap every hold at $0.15\,$s and skip the sub-point if it falls within $0.05\,$s of the preceding decision.

Placing the sub-point requires the length of the ongoing interval, which cannot be measured while it is running because the event that closes it has not yet occurred.
Intervals alternate between stance and flight, and the two differ in duration, so we estimate the ongoing interval from the most recent one of the same type, the gap between the third- and second-most-recent events.
The sub-point is placed at half the estimated interval after the most recent event.

We trigger decisions at GRF events rather than at fixed intervals so that the policy remains free to choose its own stride period.
If each command were held for a fixed interval of simulated time, the controller would be locked to that clock, and in our experience the stride period then settles at multiples of the hold, which prevents the policy from lengthening or shortening it in search of more speed.
Because GRF events occur at fixed points in the gait cycle, triggering on them keeps the decisions synchronized with gait phase while leaving the stride period under the policy's control.
This scheme relies on the cyclic structure of gait.
How decision points should be placed for tasks without such structure remains open, and we discuss possible criteria in Section~\ref{sec:disc-holding}.

\subsection{Reinforcement-learning setup}
\label{sec:rl}

The policy is trained with Soft Actor--Critic (SAC) \citep{haarnoja2018} using generalized State-Dependent Exploration (gSDE) for temporally correlated exploration noise \citep{raffin2021gsde}, implemented on top of Stable-Baselines3 \citep{raffin2021sb3}.
The same configuration is used for the $\lambda$-hold controller and for all SAC-based baselines.
The DEP-RL baseline instead combines Differential Extrinsic Plasticity (DEP), a self-organized exploration signal \citep{der2015}, with Maximum a-posteriori Policy Optimization (MPO) \citep{abdolmaleki2018mpo}, as described in Section~\ref{sec:baselines}.
Full hyperparameters for both stacks are listed in Appendix~\ref{app:hyperparams}.
Because each hold interval constitutes a single transition, the exploration noise perturbs $\lambda$ once per decision step and stays constant across the interval.

The observation is restricted to signals the nervous system can sense \citep{proske2012}.
It comprises muscle length and velocity (spindle Ia and II), muscle--tendon unit force (Golgi tendon organ Ib), foot contact and GRF, joint angles and angular velocities, and vestibular signals, together with an efference copy of the $\lambda$ currently being held, giving a total of 432 dimensions.
Excitation and activation are excluded, since they lie downstream of that command and are fully determined by $\lambda$, $\ell_i$, and $v_i$.

\subsection{Task: sprinting with a minimal reward}
\label{sec:task}

We evaluate the $\lambda$-hold controller on a sprinting task.
The per-step reward is exponential in the center-of-mass (COM) forward velocity $v_x$, attenuated by a narrow Gaussian on mediolateral velocity $v_z$:
\begin{equation}
  r \;=\; \exp(v_x)\,\cdot\,
          \exp\!\big(-(v_z/\sigma_{\mathrm{lat}})^2\big)
  \label{eq:reward}
\end{equation}
Here $\sigma_{\mathrm{lat}}=0.1\,$m/s.
The velocity term is uncapped, so the reward keeps rising with forward speed instead of saturating at a moderate speed.
The lateral term is a multiplicative factor in $[0,1]$ that discourages sideways wobble.
An episode terminates early when the COM height drops below $0.5\,$m or the head height drops below $0.9\,$m.

We chose sprinting because its objective is uncontroversial.
For walking or standing balance, for example, any reward function encodes a contestable assumption about what the nervous system optimizes.
Sprinting carries no such ambiguity, since moving forward as fast as possible is what the task means.
The reward can therefore stay minimal.
That so little suffices for human-like sprinting to emerge is itself one of our findings (Section~\ref{sec:res-sprint}).

\subsection{Baselines}
\label{sec:baselines}

We compare $\lambda$-hold against four baselines, all trained on the identical task and model.

\textbf{Plain SAC.}
The policy commands the 90 muscle excitations directly. This is the standard muscle-level action space and serves as the null baseline.

\textbf{Excitation-hold.}
The policy commands excitations directly and holds them between the same decision points as $\lambda$-hold, with no reflex. This is action repeat \citep{mnih2015,sharma2017} applied at the excitation level, included not as a method from the muscle-control literature but as an ablation that isolates intermittency on its own. Any advantage of $\lambda$-hold over it is therefore attributable to the stretch-reflex recruitment law rather than to holding alone.

\textbf{DEP-RL} \citep{schumacher2023deprl}.
The policy commands excitations directly, driven by DEP\,+\,MPO. We run the implementation in the previous study rather than re-implementing it, and keep the default hyperparameters with one deliberate adjustment. The original configuration uses a $0.025\,$s simulation step, so we rescale the step-counted quantities to preserve their physical timescales at our $0.01\,$s step, multiplying time windows and delays by $2.5$, per-step rates by $1/2.5$, and replacing the discount $\gamma$ with $\gamma^{1/2.5}$.

\textbf{Synergy} \citep{park2026}.
The policy commands a low-dimensional latent representation rather than individual muscle excitations. Following the original study, the latent has 30 dimensions. Ten dimensions encode synergy coefficients for each leg (20 dimensions in total), which a fixed non-negative matrix factorization (NMF) basis \citep{lee1999nmf}, shared across both legs, expands into the corresponding leg's muscle excitations. The remaining 10 dimensions independently control trunk-muscle excitations. We keep the synergy structure but omit the policy mirroring used during training in the original study since we are asking what the synergy action space achieves on its own and our own controller uses no symmetry aid. The original study obtained the basis by NMF of muscle-activation profiles estimated from human data. We instead obtain the basis by NMF of the $\lambda$-hold controller's own best rollouts, following the idea of extracting synergies from a trained policy \citep{berg2023sar}. This choice favors the baseline, whose basis is distilled from a strong in-simulation solution rather than transferred from human recordings. The comparison therefore addresses whether $\lambda$-hold still outperforms the best fixed basis extracted from its own behavior.

\subsection{Quantifying learning speed: simulation steps and decision steps}
\label{sec:learning-speed}

The controllers differ in what the policy commands and in how often it is queried, so the number of simulation steps and the number of policy queries are not interchangeable measures of training cost.
We therefore report the performance along two axes.
A simulation step is one engine step of $0.01\,$s (Section~\ref{sec:model}), and a decision step is one policy query.
Plain SAC and Synergy query the policy at every simulation step, so for them the two axes coincide.
Excitation-hold and $\lambda$-hold controllers hold one command across many simulation steps, so their decision steps are far fewer.
In DEP-RL the exploration rule intermittently takes over from the policy during data collection, but it does so rarely, so its decision count falls only slightly below its simulation step count.

The two axes answer different questions.
The simulation-step axis measures performance per unit of environment experience, putting all methods on equal footing.
The decision-step axis measures performance per policy query, which determines the computational cost of training and, at deployment, how often the controller must run.

\subsection{Joint-velocity coverage}
\label{sec:coverage}
To measure exploration directly, we quantify the variety of movements the agent produces over training.
Joint angles describe posture rather than movement, so we use the model's joint angular velocities.
Of the 21 degrees of freedom, we exclude the six of the pelvis root, which describe global body motion rather than joint coordination, leaving the 15 anatomical joint velocities at the hip, knee, ankle, subtalar, and lumbar joints.

We report two measures against the simulation-step count.
The first is the entropy of the visited velocity distribution, which captures how varied the movements are.
We estimate it with the Kozachenko--Leonenko $k$-nearest-neighbor estimator \citep{kozachenko1987sample}, which infers differential entropy from the distance of each sample to its $k$-th nearest neighbor ($k=4$), applied to velocities standardized per dimension within each method.
The estimate depends on sample size, so we draw the same number of samples from every method.

The second is grid coverage, which measures how much of the velocity space is explored.
The bin edges are fixed once using pooled data from all methods, with each method contributing an equal number of samples.
Each dimension is divided into four equal-width bins spanning its 1st to 99th percentile, excluding extreme outliers.
The resulting grid contains $4^{15}\approx1.1\times10^{9}$ cells.
Coverage at a given point in training is the number of distinct cells visited up to that point.
No method reaches more than a small fraction of the total grid.

\subsection{Validation of human-likeness}
\label{sec:human-eval}
We assess how human-like the emergent gait is along three axes, sagittal joint kinematics, vertical GRF, and muscle activity, each compared against openly available human running data.

The three comparisons share the same processing.
Cycles are aligned at right initial contact.
The kinematics and GRF dataset is supplied already normalized to the gait cycle, the human EMG recordings are segmented at marker-based heel-strike \citep{zeni2008}, and the model is segmented at the rising edge of the right vertical GRF.
Each cycle is time-normalized to 101 points spanning 0 to 100\% of the cycle, and the cycles are averaged into a single mean waveform.
Agreement is reported per variable as the Pearson correlation between the model and human mean waveforms.

For kinematics and the vertical GRF we use the treadmill-running dataset reported by Fukuchi et al.\ \citep{fukuchi2017} at $2.5$, $3.5$, and $4.5\,$m/s.
The emergent sprint accelerates from a standing start, so we group the model's gait cycles by mean COM speed, keeping the cycles within $\pm 0.35\,$m/s of each reference speed.
We then compare each group with the human data recorded at the corresponding speed.
The lower speed bins therefore represent accelerating gaits rather than steady state gaits, whereas the $4.5\,$m/s bin lies nearest to the model's top speed, enabling more direct comparison between steady-state gaits of the model and human.

For muscle activity, we compare the model's activations with the surface EMG reported in \citep{vanhooren2024}, recorded from 19 participants running at $5.0\,$m/s.
This is the fastest openly available running EMG dataset and therefore the closest to the model's top speed.
We take all model cycles with a mean COM speed of at least $4.3\,$m/s.
The human EMG is band-pass filtered at 20 to 450\,Hz, full-wave rectified, and low-pass filtered at 10\,Hz to give linear envelopes.
The envelope corresponds to activation rather than excitation, since its low-pass stage plays the same role as the activation dynamics of the muscle model.
Each waveform is peak-normalized to its own maximum for display.
We report the correlation for each of the 11 muscles and their mean.

\section{Results}
\label{sec:results}

\subsection{\texorpdfstring{$\lambda$}{λ}-hold learns orders of magnitude faster}
\label{sec:res-learning}

\begin{figure}[tb]
  \centering
  \includegraphics[width=\textwidth]{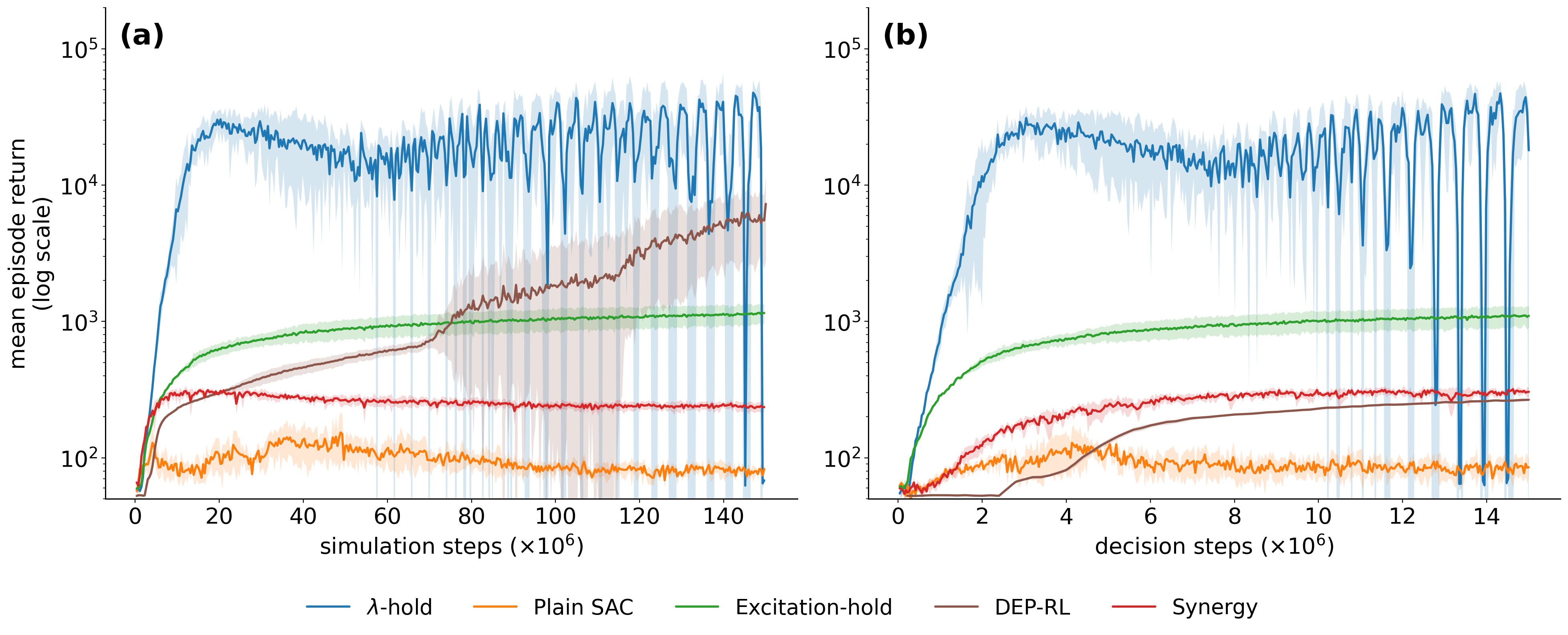}
  \caption{Sprint learning curves for $\lambda$-hold and the four baselines, trained under an identical simulation-step budget. Each solid line is the mean over five random seeds and the shaded band shows $\pm$ standard deviation. The return axis is logarithmic, so the plotted gap understates how far $\lambda$-hold outperforms the baselines. The deep, regular downward spikes in the $\lambda$-hold curve, visible on both axes, are the late-training oscillation analyzed in Appendix~\ref{sec:supp-rsi}.
  (a) Return against simulation steps, the axis on which all methods receive equal environment experience.
  (b) Return against decision steps, the number of policy queries.}
  \label{fig:learning-curves}
\end{figure}

Figure~\ref{fig:learning-curves} shows the mean episode return on axes of simulation steps and decision steps.
On the simulation-step axis, which gives every method equal environment experience, $\lambda$-hold separates from every baseline within the first $10\times10^{6}$ steps and stays above them for the rest of training.
The emergent gaits for all methods can be viewed at \url{https://lee-jun-hyuk-37.github.io/projects/lambda-hold/}.

The baselines fall into two groups.
Plain SAC and Synergy do not progress beyond a rudimentary form of locomotion.
They take one or two steps, kick the stance foot hard backward, pitch the trunk forward, and dive to the ground, after which training stalls.
This explains why prior muscle-level work has required reference-motion imitation, reward shaping, hand-designed control rules with tuned gains, or curricula to achieve convergence at all (Section~\ref{sec:prior}).
DEP-RL improves slowly but steadily and overtakes the other baselines late in training.
By the end, it learns to balance and move forward without falling, at about $2.6\,$m/s.
However, its gait is not human-like: the trunk leans far forward, and the gait is markedly asymmetric.
It resembles an awkward fast walk rather than a sprint.

Excitation-hold isolates the contribution of holding.
It commands excitations directly, as Plain SAC does, but holds them between the same decision points as $\lambda$-hold, and it settles about an order of magnitude above Plain SAC.
Action repeat is therefore substantially beneficial at the muscle level.
Nonetheless, the episode return by Excitation-hold is left more than an order of magnitude below that of $\lambda$-hold; the larger part of the advantage of the $\lambda$-hold controller comes from changing the control variable to $\lambda$ rather than from holding.

On the decision-step axis, $\lambda$-hold and Excitation-hold reach a given return with about an order of magnitude fewer policy queries than Plain SAC, Synergy, and DEP-RL, which are queried at every simulation step.

On both axes the $\lambda$-hold curve oscillates late in training, as the policy periodically loses the fast sprint, the model falls, and the sprint is then relearned.
This comes from learning the acceleration and the high-speed phases at once, and is analyzed with its remedy in Appendix~\ref{sec:supp-rsi}.

Holding $\lambda$ also makes training inexpensive in absolute terms.
On our local workstation, a single NVIDIA RTX~5070 GPU with a 24-thread Intel Core~Ultra~9 CPU, a sprint reaching about $4.0\,$m/s emerges in roughly an hour of wall-clock training, which makes it practical to run large batches of hypotheses.
We do not report a head-to-head wall-clock comparison across methods because the simulation-to-decision ratio, the optimizer, the parallel-environment configuration, and the simulator throughput all differ.
Any such factor, rather than the control scheme itself, would be partly responsible for the resultant differences.

\subsection{\texorpdfstring{$\lambda$}{λ}-hold explores joint space far more widely}
\label{sec:res-coverage}

\begin{figure}[tb]
  \centering
  \includegraphics[width=\textwidth]{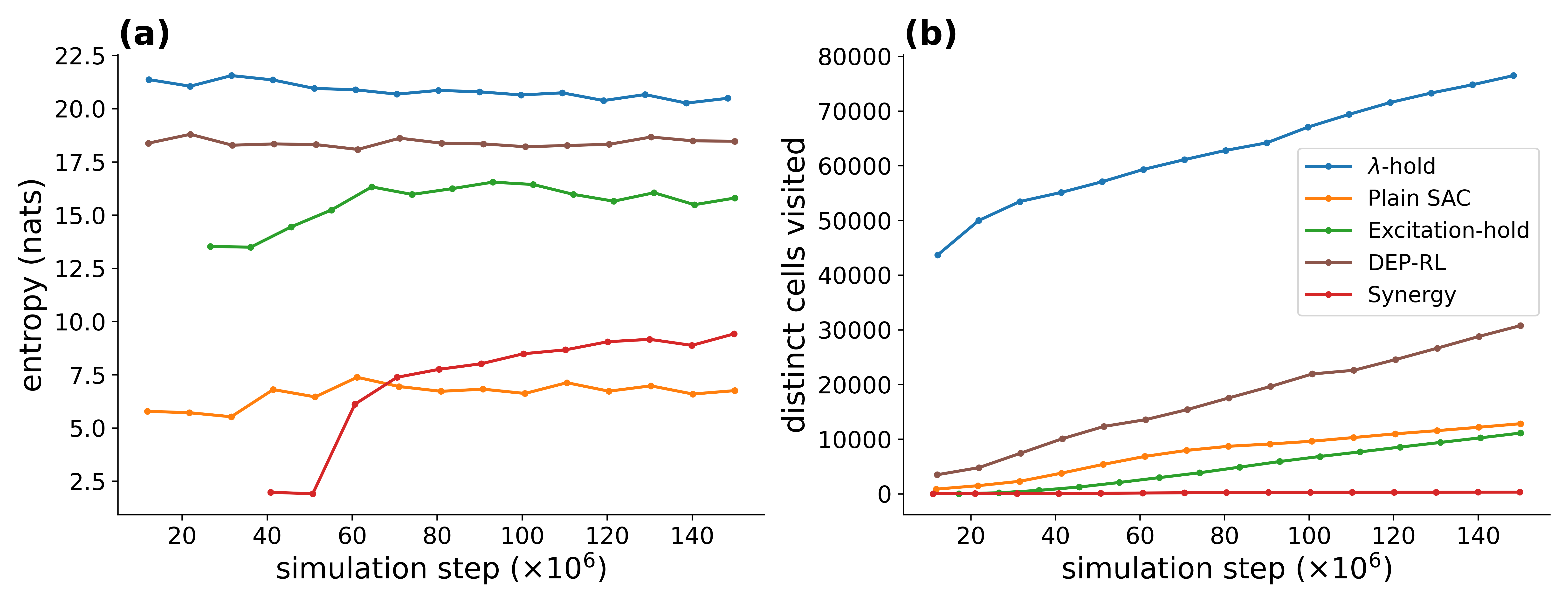}
  \caption{How widely each controller explores joint-velocity space over training.
  (a) The entropy of the visited joint-velocity distribution, a measure of how varied the movements are, estimated with a Kozachenko--Leonenko $k$-nearest-neighbor estimator at an equal sample size ($n=150$) across methods. Excitation-hold and Synergy have no estimate at the earliest points because they had not yet accumulated that sample.
  (b) The coverage of joint-velocity space, a measure of how much of the movement space is reached, taken as the cumulative number of distinct cells visited on a grid shared across methods, with 4 bins along each of the 15 joint-velocity dimensions.}
  \label{fig:coverage}
\end{figure}

Figure~\ref{fig:coverage} reports the two exploration measures against the simulation-step count.
The $\lambda$-hold control yields the highest entropy from the first evaluation point and holds it roughly constant thereafter, so its movements are the most varied at every stage of training.
It also visits the most distinct joint-velocity cells throughout, finishing with about two and a half times as many as DEP-RL, the strongest baseline on this measure, and about seven times as many as Plain SAC and Excitation-hold.

Synergy finishes with the lowest coverage yet has higher entropy than Plain SAC.
Because Synergy forces synergist muscles to co-vary, joint-space movement appears readily once the policy has learned anything, which lifts the entropy above that of independent per-muscle excitation.
On the other hand, that pattern of co-variation is fixed, so the movements it can produce are confined to what the basis spans, which explains why the coverage remains the lowest of any method.

\subsection{Human-like sprinting emerges from a minimal reward}
\label{sec:res-sprint}

Under $\lambda$-hold control with the minimal reward of Eq.~\eqref{eq:reward}, coordinated sprinting emerges with strong left--right alternation and clear swing-limb flexion. The symmetry is not built into the controller. The policy sets each muscle's threshold independently, with no mirroring and no symmetry reward, so a symmetric gait is what the reward selects rather than what the architecture imposes. The resultant sprint reaches about $4.7\,$m/s at its fastest.

\begin{figure}[tb]
  \centering
  \includegraphics[width=\textwidth]{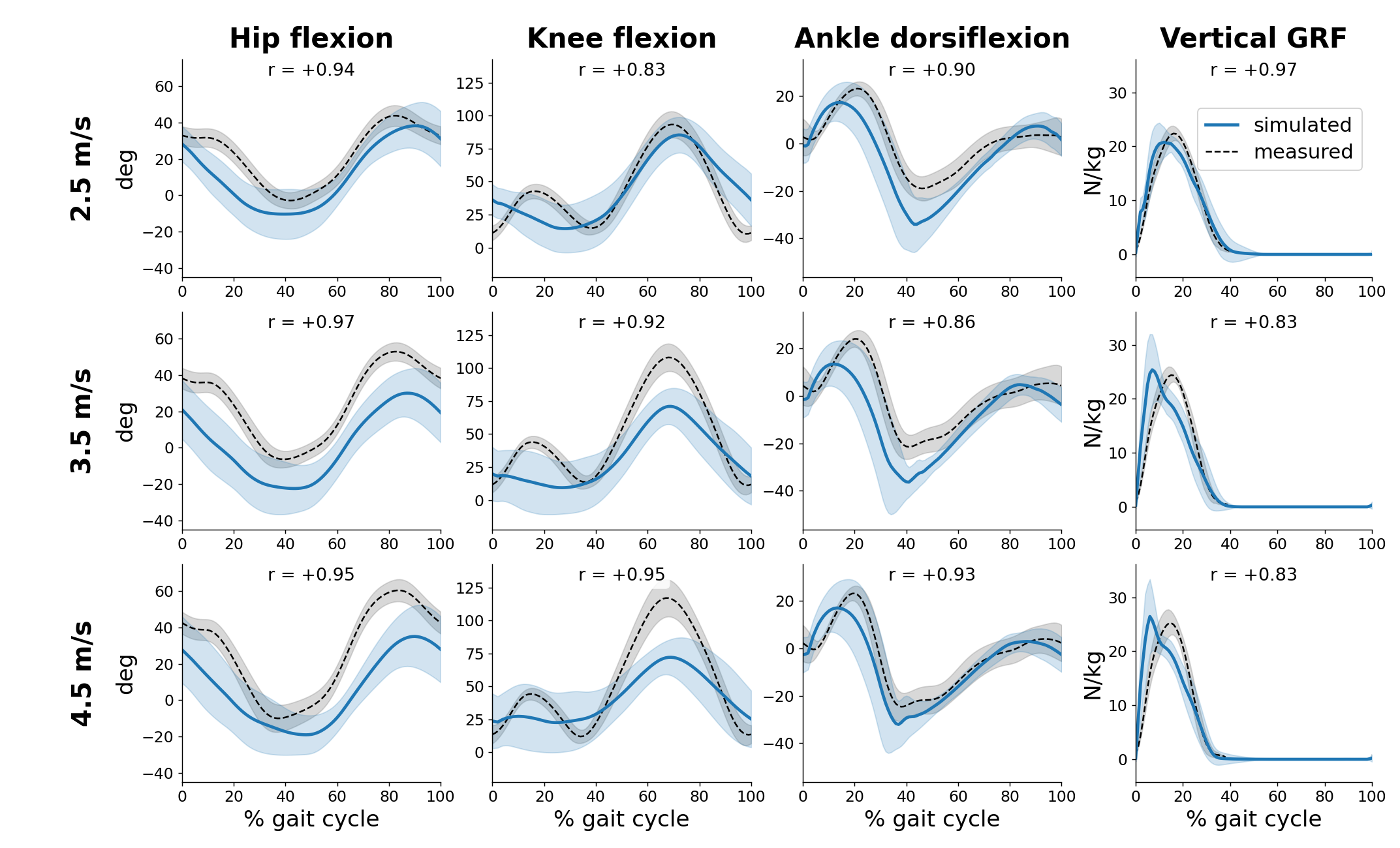}
  \caption{Human-likeness of the emergent sprint, speed-matched to human treadmill running \citep{fukuchi2017}. Simulated gait cycles are grouped by their mean COM speed, keeping those within $\pm 0.35\,$m/s of each reference speed ($2.5$, $3.5$, and $4.5\,$m/s). Both bands are mean $\pm$ standard deviation. Each simulated band pools $32$, $54$, and $218$ gait cycles at $2.5$, $3.5$, and $4.5\,$m/s, and each measured band pools $31$, $39$, and $31$ runners.}
  \label{fig:sprint}
\end{figure}

The model largely reproduces the shape of the waveforms observed in kinematics of human sprinting, and the vertical GRF with a single stance pulse and human-like magnitude at every speed (Figure~\ref{fig:sprint}). The sagittal angles depart from the human data in amplitude, and the gap widens with speed. At $2.5\,$m/s the knee swing-flexion peak nearly matches the human ($85^\circ$ vs.\ $93^\circ$), but the human knee flexes progressively more as speed rises ($93^\circ\!\to\!108^\circ\!\to\!117^\circ$) whereas the model's does not. Hip flexion follows the same pattern; the peak hip flexion falls from within the human band at $2.5\,$m/s to values below human data at higher speed conditions. We take up this departure, together with the model's speed ceiling, in Section~\ref{sec:disc-topspeed}.

\subsection{Muscle activation is moderately consistent with measured electromyography}
\label{sec:res-emg}

\begin{figure}[tb]
  \centering
  \includegraphics[width=\textwidth]{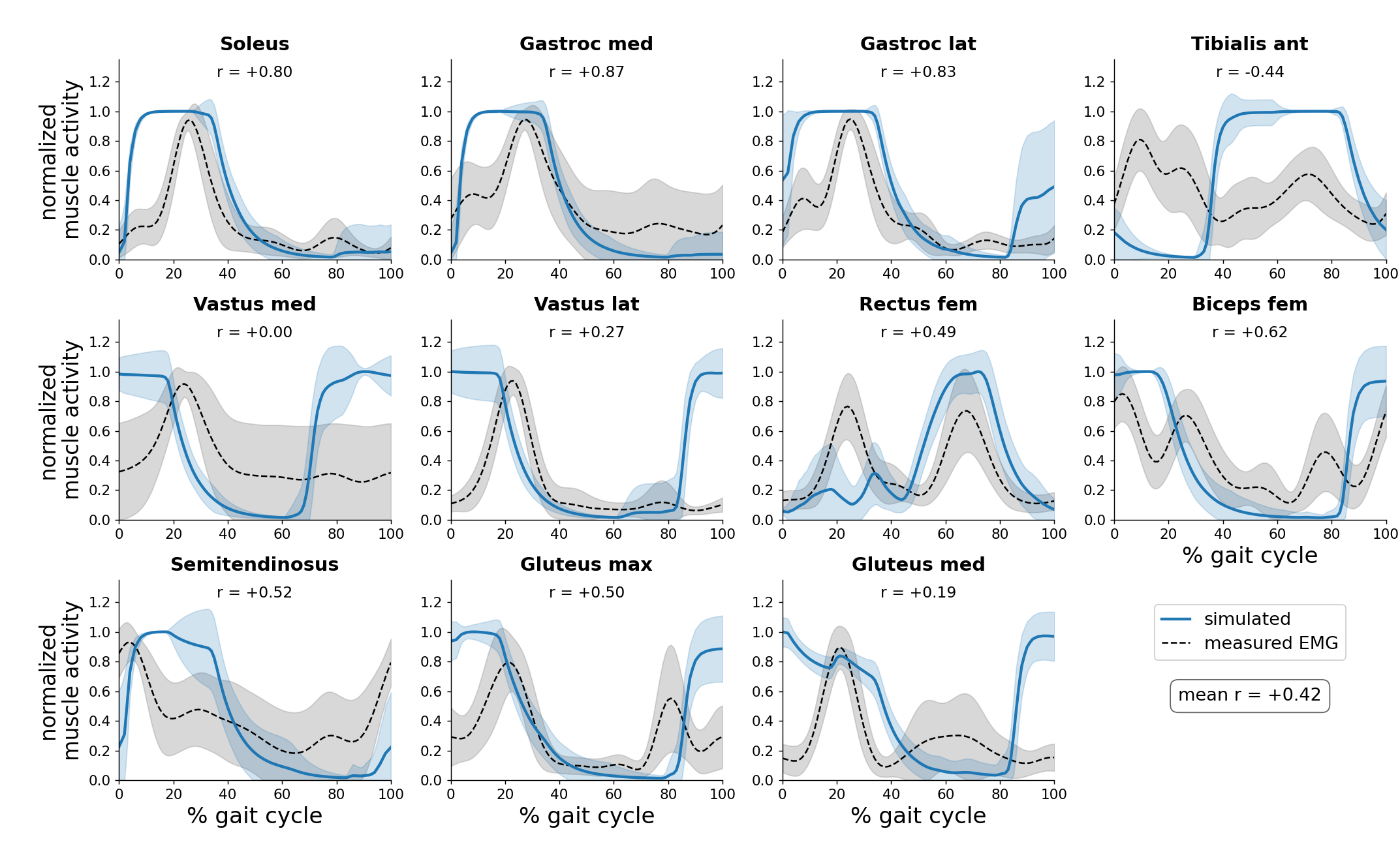}
  \caption{Model muscle activation against measured surface EMG. Model activations (blue) pool all $129$ gait cycles whose mean COM speed exceeds $4.3\,$m/s, aligned at right initial contact. Measured EMG (gray) is the mean $\pm$ standard deviation of $19$ subjects running at $5.0\,$m/s \citep{vanhooren2024}. Both are peak-normalized over the gait cycle.}
  \label{fig:emg}
\end{figure}

Human-like kinematics and kinetics do not guarantee that the underlying muscle activity is human-like, so we also compare the model's activations against measured surface EMG (Figure~\ref{fig:emg}).
The human recordings are at $5.0\,$m/s, slightly above the model's top speed of $4.7\,$m/s.

The activations agree moderately with EMG, with a mean waveform correlation of $r=+0.42$.
Agreement is strong for the propulsive plantarflexors (soleus $0.80$, gastrocnemius medialis $0.87$ and lateralis $0.83$) and moderate for the hamstrings (biceps femoris $0.62$, semitendinosus $0.52$), rectus femoris ($0.49$), and gluteus maximus ($0.50$).
It is poor for gluteus medius ($0.19$), vastus medialis and lateralis ($0.00$ and $0.27$), and the tibialis anterior ($-0.44$).
The disagreement is concentrated in swing.
The model activates the gluteus medius, vastus medialis and lateralis through late swing, where the human activity is low, and it fires the tibialis anterior through swing instead of just after contact.

This level of agreement should be interpreted in the context of the comparison and the evaluation criteria in the related field.
The controller is trained from scratch, without any experimental data.
The correlation is computed on peak-normalized waveforms, meaning that it evaluates the timing and shape of muscle activation rather than the absolute magnitude; this is the way by which models of this kind are typically evaluated \citep{hicks2015}.
Six of eleven muscles achieve $r>0.5$. Static optimization and computed muscle control, which receive the measured trajectory as input, achieve $r>0.5$ in only 6 and 11 of 40 muscle-by-speed conditions, respectively, during walking \citep{trinler2018}. The pattern of disagreement is similar. In that study, agreement was substantially better for the shank muscles than for the thigh muscles, as also observed in this study.

\section{Discussion}
\label{sec:discussion}

\subsection{Substantially more efficient exploration through \texorpdfstring{$\lambda$}{λ}-hold control}
\label{sec:disc-coverage}

The joint-velocity result (Section~\ref{sec:res-coverage}) shows where the advantage of $\lambda$-hold originates. Exploration at the excitation level in previous approaches fails because per-muscle perturbations largely cancel in the many-to-one excitation-to-joint map, so the agent moves in excitation space without moving in joint space. Perturbations in $\lambda$-space instead move the muscles in a mechanically coordinated way, and they transfer into joint space. The $\lambda$-hold therefore attains the highest entropy from the first evaluation point and accumulates coverage far faster than any baseline. A learner that visits a broader range of states sooner also finds high-performing states sooner, which is the proximate reason for the faster return growth.

The coordination has a simple mechanical origin. Muscles that act as synergists share mechanical action, so their lengths and velocities co-vary throughout a movement almost by definition. Under the reflex law of Eq.~\eqref{eq:reflex}, each muscle's excitation is determined by its own length and velocity relative to the held threshold, so synergist muscles are necessarily recruited together. A $\lambda$-hold controller therefore produces synergy-like recruitment without ever being given a synergy basis. Crucially, this coordination is not a fixed matrix, as in synergy models built on a factorization of measured muscle activity \citep{davella2003,chvatal2013}. Because the coupling runs through the mechanical state, it changes with posture and with the phase of the movement, like the task- and state-dependent synergies observed in humans. The Synergy baseline shows the limitation of a fixed basis, spreading its movements broadly relative to its own range while that range stays the narrowest of any method. The $\lambda$-hold, in contrast, gives the learner the coordinated exploration that synergy action spaces were introduced to provide, without committing it to a single fixed pattern.

Holding the command helps in a second way. A perturbation applied to $\lambda$ persists across the whole interval, so it produces a sustained deviation in the movement rather than jitter that averages out within a stride. Fewer decisions per unit of simulated time also shorten the chain over which the learner must assign credit. Excitation-hold isolates this effect. It differs from Plain SAC only in holding its commands, and it reaches about an order of magnitude higher return (Section~\ref{sec:res-learning}). Holding is also consistent with how humans are thought to control movement, which we take up in Section~\ref{sec:disc-holding}.

\subsection{Implementing principles of human motor control in predictive simulation}
\label{sec:disc-principles}

When implementing predictive simulation using a muscle-driven skeletal model, the fidelity of the musculoskeletal model is critical to ensuring the reliability of the simulation. Accordingly, many simulation studies have adopted Hill's muscle model \citep{hill1938,zajac1989,millard2013}, whose fidelity has been extensively validated through previous research in physiology, and have used physiologically faithful model parameters \citep{delp2007opensim,arnold2010,rajagopal2016}. In this approach, the range of possible simulations is constrained not only by the mechanics of the musculoskeletal system but also by the constraints imposed by muscle physiology.

We propose that, just as fidelity to muscle physiology is important, fidelity to human motor control is also necessary for a valid predictive simulation. In other words, predictive simulations should incorporate not only the constraints imposed by muscle physiology but also the constraints arising from the principles of motor control that humans are thought to employ to effectively coordinate multiple muscles. To this end, just as Hill's muscle model provides an effective means of faithfully representing muscle physiology, incorporating principles of human motor control whose validity has been supported by previous studies may provide an effective means of faithfully representing motor control and the resulting behavior. In this study, we adopted the $\lambda$ version of EP hypothesis and the mechanism of intermittent control.

\subsubsection{The EP hypothesis, its objections, and reframing through learned policy}
\label{sec:disc-reframe}

Feldman proposed that muscle forces and the resulting equilibrium emerge from the interaction between referent thresholds and the mechanical properties of the body \citep{feldman1986}. This EP hypothesis has since received support from a range of experimental and theoretical studies \citep{feldmanlevin1995,latash2010synergies}. At the same time, as with many influential hypotheses in human motor control, the EP hypothesis has also been challenged. One recurring objection is that fast or complex movements appear to require virtual trajectories that are too elaborate to be physiologically plausible. This has been taken as evidence that the central nervous system must instead compute an explicit internal model of limb dynamics \citep{gomi1996}.

However, sprinting, which is generated by the predictive model in this study, does not fall within the scope of this first objection. Although sprinting is both fast and highly multi-joint, the referent commands in our controller are updated only eight times per gait cycle and remain constant between updates. Moreover, the same reflex law, with the same gains, is applied across all 90 muscles. Thus, the production of sprinting in this model does not require a finely shaped or continuously updated virtual trajectory. At least for this class of behavior, the complexity of the observed movement need not imply a correspondingly complex sequence of referent commands.

A second, more recent objection concerns a different limitation of EP hypothesis: the framework specifies how referent commands can generate movement but does not by itself explain where those commands come from. In realistic multi-joint behavior, this can leave the theory with what has been described as a computational homunculus \citep{mangalam2026}. In our controller, the source of the referent commands is a learned policy. This policy does not constitute a homunculus in the sense of an agent that is simply assumed to have already solved the control problem; rather, it is constructed through a specified learning process driven by a reward function. Nevertheless, this implementation does not eliminate the possibility of an internal model so much as relocate it: if an internal model is required, it may be embedded in the learned policy rather than in the low-level threshold controller itself. Importantly, what the policy learns is not directly interpretable from the resulting controller. We therefore present this architecture as a modeling demonstration of how threshold control can be embedded within a control system, rather than as a claim about the actual organization or computational implementation of the human nervous system.

Our design also speaks to a broader debate concerning the relationship between the EP hypothesis and optimal feedback control, a leading alternative account of motor coordination, in which the nervous system is characterized as an optimizer that generates corrective actions on the basis of an estimate of the body's state \citep{todorov2002}. From a traditional EP perspective, these frameworks have been viewed as competing accounts, partly because computational approaches of this kind appear to require the nervous system to represent the forces or dynamics needed to produce a movement---precisely the type of explicit force computation that EP hypothesis was formulated to avoid \citep{ostry2003}.

In our model, these two principles need not be mutually exclusive. The learned policy in our architecture is itself a state-feedback controller, but its optimization operates over referent commands rather than directly over muscle forces. The EP-based reflex mechanism then translates those referents into muscle activation, while the resulting forces emerge from the interaction between the reflex dynamics and the mechanical properties of the body. From this perspective, optimal feedback control and EP control can be understood not as competing explanations, but as descriptions of different levels within a hierarchical control architecture: optimization operates at the higher level to determine appropriate referents, while the EP-based reflex mechanism provides the lower-level substrate through which those referents are realized mechanically. This interpretation does not establish that the nervous system is organized in this manner, but it demonstrates that optimization over behavioral goals and threshold-based emergence of muscle forces can coexist within a single computational framework.

\subsubsection{Intermittent control and event-driven command updating}
\label{sec:disc-holding}

The idea that motor commands are issued intermittently rather than continuously has also been proposed in human motor control. The intermittent-control framework argues that motor commands are generated as sparse, discrete updates, while muscle viscoelasticity and spinal reflexes maintain control between updates---an arrangement described as continuous observation with intermittent action \citep{gawthrop2011}. The $\lambda$-hold controller has a closely related structure. The reflex mechanism continuously reads muscle length and velocity at every simulation step, whereas the policy intervenes only at discrete decision points. In our implementation, these updates occur more frequently than the intervals reported in studies of human motor control \citep{loram2014,vandekamp2013}, but the underlying distinction between continuous sensing and intermittent command updates is the same.

The two traditions---EP-based threshold control and intermittent control---however, address complementary aspects of control. EP hypothesis specifies the variable to be controlled---the referent threshold---but does not prescribe how frequently that threshold should be updated \citep{feldman1986,feldmanlevin1995,latash2010synergies}. Intermittent-control theory, in contrast, specifies the temporal structure of command updating but does not, by itself, determine which variable should be controlled \citep{gawthrop2011}. The $\lambda$-hold controller combines these two principles: the controlled variable is a 90-dimensional vector of muscle-specific thresholds, which is updated at discrete mechanical events and held constant between them. Importantly, intermittent updating is not foreign to the EP framework itself. The EP account treats changes in the control command as discrete events \citep{feldman2019}, and experimental work has shown that referent shifts can be completed well before the movement they generate has come to an end \citep{ghafouri2001}. To our knowledge, however, these ideas have not previously been combined into a muscle-level controller in which referent thresholds are explicitly held between discrete updates.

This leaves an important question: what determines when a new command should be issued? Our GRF-based trigger (Section~\ref{sec:hold-timing}) provides one concrete implementation of event-driven intermittent control. It is particularly suitable for locomotion because salient gait events provide natural points at which the controller can reconsider its command. The same problem has a close analog in reinforcement learning, where the timing of a new decision can be determined by a termination condition that is itself learned \citep{bacon2017} or by a cost associated with deliberation at each decision point \citep{harb2018}. In motor control, another possibility is to trigger an update when prediction error becomes sufficiently large: a new command would be issued only when the current state departs substantially from the state expected under the held command. This predictor-based formulation is consistent with the prediction-error account of intermittent control proposed by \citet{gawthrop2011}.

The GRF trigger should therefore be viewed as one task-specific solution rather than as a general account of command timing. A genuinely task-agnostic timing mechanism would need to determine when the current command is no longer adequate in relation to the agent's task or intent. Whether such a mechanism can be learned from experience, derived from prediction error, or grounded in some other principle remains an open question. Importantly, separating the questions of what is controlled and when the control command is updated would provide a useful way to connect EP-based threshold control with the broader literature on intermittent and event-driven motor control.

\subsection{Why the emergent sprint falls short of human top speed}
\label{sec:disc-topspeed}

The emergent sprint is fast and human-like in coordination but does not reach the top speed of elite human sprinters, for two reasons that lie in the model rather than in the control scheme. The first is the muscle force--velocity relation. During the concentric swing drive the hip flexors are fully activated, yet they shorten fast enough that the force--velocity relation cuts their force to between a quarter and a third of its isometric value (Appendix~\ref{sec:supp-fv}). Full activation therefore buys little propulsive force in the phase that would accelerate the limb. This is the likely source of the reduced swing-flexion amplitudes shown in Section~\ref{sec:res-sprint}. A simulation study has identified this relation as the dominant contractile limit on maximum sprinting speed \citep{miller2012}, and relieving it in our own model raises the top speed from about $4.7$ to $6.5\,$m/s. The second is that our model has no arms. During running the upper extremities generate angular momentum that offsets the angular momentum of the swinging legs and helps regulate whole-body rotation \citep{hinrichs1987}, so their absence removes a contributor to balanced high-speed gait.

\subsection{Toward a learnable model of the motor controller}

The most immediate next step is to test whether the substrate generalizes beyond sprinting to balance, jumping, and walking. These would also test the hold-timing question in regimes without clear cyclic events. Longer term, a genuine digital clone of the human motor controller must reproduce human-like responses to perturbation, and must generalize across conditions before it can be personalized to an individual, including to the altered strategies of patients and older adults. Those are the clinical applications that motivate a learnable model of the motor controller in the first place.

\section*{Acknowledgments}
This work was supported in part by the Korea Health Technology R\&D Project through the Korea Health Industry Development Institute (KHIDI) funded by the Ministry of Health \& Welfare (No.\ HK23C0071) and the National Research Foundation of Korea (NRF) grant funded by the Korea government (MSIT and MOE) (No.\ RS-2026-25500579).

\section*{Code availability}
The code used in this study is available at \url{https://github.com/Lee-Jun-Hyuk-37/Lambda-Hold} and archived at Zenodo (\url{https://doi.org/10.5281/zenodo.23265020})\citep{lambdahold_zenodo} under the MIT license. The environments were built on sconegym (\url{https://github.com/tgeijten/sconegym}, commit db2d2c1). Simulations were performed in SCONE (version 2.4.3.3209) with the Hyfydy simulation engine (version 1.10.6.1358), which requires a license (a free trial license is available at \url{https://hyfydy.com}).

\bibliographystyle{unsrtnat}
\bibliography{references}

\appendix
\section*{Appendix}

\section{Training hyperparameters}
\label{app:hyperparams}

Table~\ref{tab:hyperparams} lists the full configuration for both learning stacks.

\begin{table}[!ht]
  \centering
  \small
  \caption{Training hyperparameters. The SAC-based column applies to the $\lambda$-hold controller and to the Plain SAC, Excitation-hold, and Synergy baselines. The DEP-RL column gives the DEP\,+\,MPO settings. DEP's step-counted quantities are rescaled from the reference $0.025\,$s step to the shared $0.01\,$s step (Section~\ref{sec:baselines}).}
  \label{tab:hyperparams}
  \setlength{\tabcolsep}{4pt}
  \hyphenpenalty=10000\exhyphenpenalty=10000
  \begin{tabular}{@{}>{\raggedright\arraybackslash}p{4.6cm} >{\raggedright\arraybackslash}p{5.5cm} >{\raggedright\arraybackslash}p{5.2cm}@{}}
    \toprule
    Setting & SAC-based & DEP-RL \\
    \midrule
    Algorithm & SAC & DEP\,+\,MPO \\
    Actor / critic network & MLP, two hidden layers of 256 units & MLP, two hidden layers of 1024 units \\
    Exploration & gSDE, resampled every 4 decision steps & DEP (defaults, rescaled to $0.01\,$s) \\
    Learning rate & $3\times10^{-4}$ & actor $3.53\times10^{-5}$, critic $6.08\times10^{-5}$, dual $2.13\times10^{-3}$ \\
    Discount $\gamma$ / target smoothing $\tau$ & $0.99$ / $0.005$ & $\gamma^{1/2.5}\!\approx\!0.996$ / --- \\
    Batch size / replay buffer & $256$ / $5\times10^{5}$ & $256$ / $1\times10^{6}$ \\
    Gradient steps / train freq & $2$ / $1$ & $30$ per update (every $2500$ steps) \\
    Entropy coefficient & auto & n/a (MPO uses KL duals) \\
    Parallel environments & $24$ & $12$ \\
    Observation / reward normalization & observation and reward (online mean and variance) & observation only (online mean and variance) \\
    \bottomrule
  \end{tabular}
\end{table}

\section{Extending the sprint: stabilizing training and raising the speed ceiling}
\label{sec:supp-rsi}

The main text characterizes the $\lambda$-hold controller on its own.
Two further ingredients extend it, one stabilizing the late-training return and the other raising the top speed. Neither is part of the controller evaluated in the main text, so we report them separately here.

\subsection{Reference-state initialization stabilizes the late-training oscillation}
\label{sec:supp-rsi-fix}

\begin{figure}[tb]
  \centering
  \includegraphics[width=\textwidth]{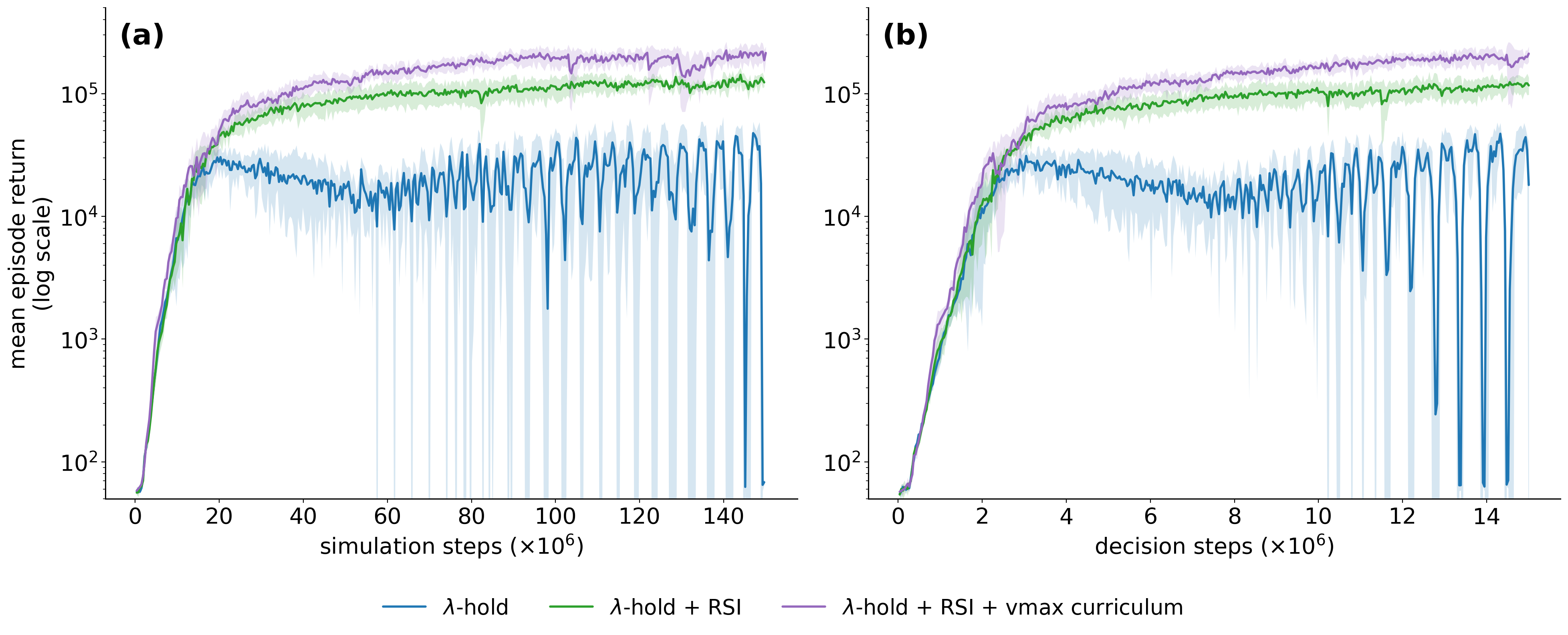}
  \caption{The effect of reference-state initialization (RSI) and the maximum-shortening-velocity curriculum.
  Three conditions are compared: $\lambda$-hold alone (blue), $\lambda$-hold with RSI (green), and $\lambda$-hold with RSI and the shortening-velocity curriculum (purple).
  (a) Return against simulation steps and (b) against decision steps.
  All three use five seeds, with the solid line the mean and the shaded band showing $\pm$ standard deviation.}
  \label{fig:rsi}
\end{figure}

The $\lambda$-hold return climbs quickly but then oscillates late in training (Figure~\ref{fig:learning-curves}), with the policy repeatedly losing the fast sprint, falling, and recovering it.
The cause is the task rather than the controller.
Sprinting from a standing start bundles two very different sub-tasks into a single episode, an initial acceleration phase from low speed and a steady high-speed running phase.
One policy that must serve both tends to forget the fast sprint while it improves the acceleration, and the reverse, which is the familiar difficulty of multi-task RL.
The oscillation is therefore a property of learning both sub-tasks at once.
It is absent from the baselines simply because they never reach the fast sprint.

To keep the fast sprint from being forgotten we add reference-state initialization (RSI) \citep{peng2018deepmimic}, under which an episode is occasionally reset not to the default standing posture but to a state the agent has already reached during the fast sprint.
Following the spirit of Go-Explore \citep{ecoffet2021goexplore}, we maintain a self-generated archive of the highest-return episodes seen so far, and after a warm-up of $3\times10^{6}$ decision steps a fraction of episode resets draw their initial state from this archive.
Revisiting these states lets the agent keep practicing the fast sprint while the acceleration phase is relearned.
With RSI the late-training oscillation disappears and the return becomes smooth and sustained (Figure~\ref{fig:rsi}).

\FloatBarrier
\subsection{The speed ceiling is the muscle force--velocity relation}
\label{sec:supp-fv}
\begin{figure}[tb]
  \centering
  \includegraphics[width=\textwidth]{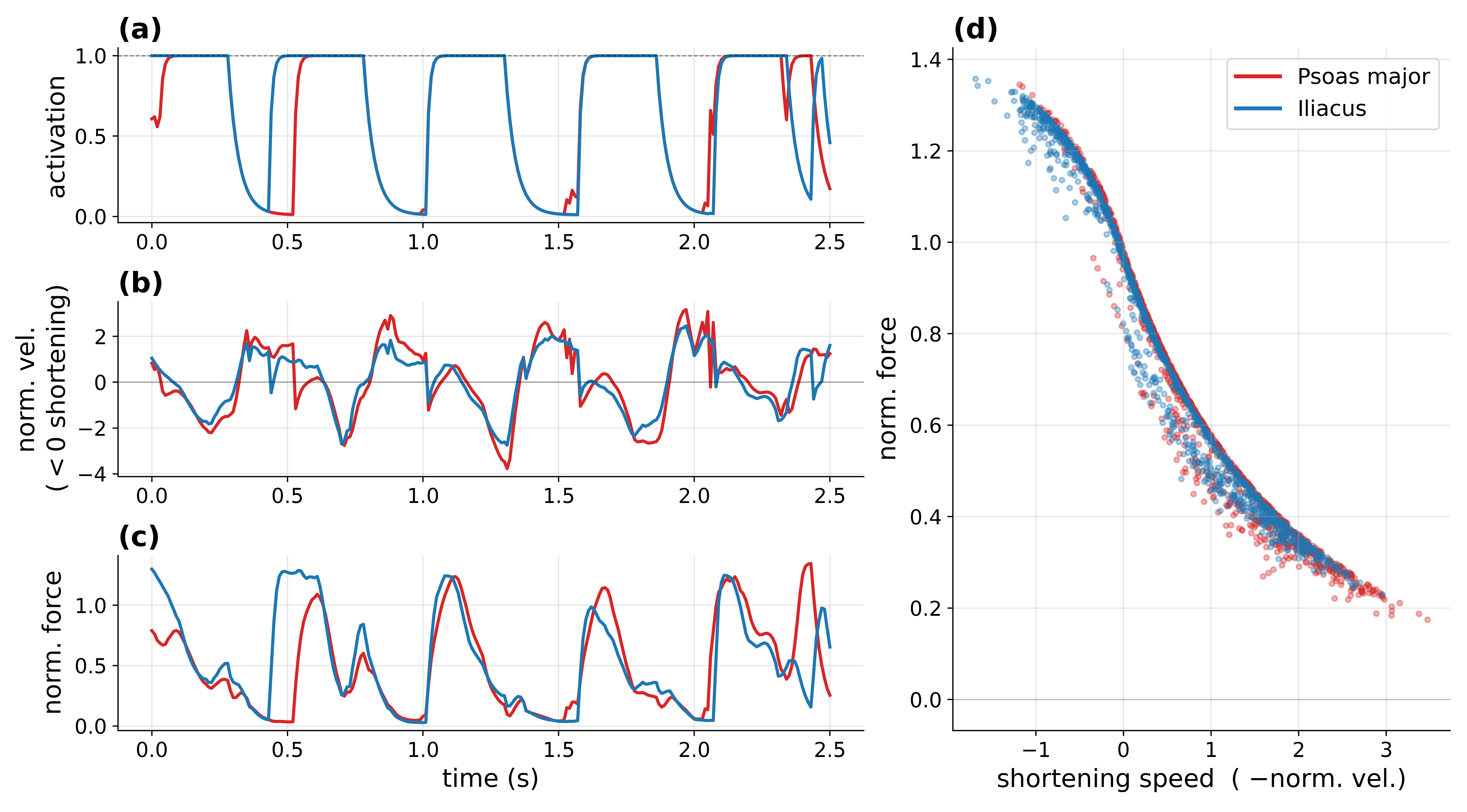}
  \caption{Hip-flexor force and velocity of the right psoas major and iliacus at the top-speed $\lambda$-hold episode over the last few gait cycles. (a) Muscle activation, which saturates at $1$ during the concentric swing drive. (b) Normalized velocity over the same window, reaching $-2$ to $-3$ (negative is shortening) just as activation saturates. (c) Normalized force, strongly modulated despite the near-constant activation. (d) Fully activated samples only (activation ${>}0.9$), plotted against shortening speed. Averaged over these instants, force is $0.95$ (psoas major) and $0.94$ (iliacus) near isometric but only $0.26$ and $0.33$ at the fastest shortening.}
  \label{fig:hipfv}
\end{figure}

Even stabilized, the controller tops out near $4.7\,$m/s, and its fastest episodes show why.
During the concentric swing drive the hip flexors are fully activated, yet because they are shortening rapidly the intrinsic force--velocity relation cuts their force to between a quarter and a third of its isometric value (Figure~\ref{fig:hipfv}).
Full activation thus yields little propulsive force in the phase that would accelerate the limb.
The ceiling is therefore a force--velocity property of the H2190 plant, not of the $\lambda$-hold controller.

To confirm this we add to RSI a curriculum on the maximum shortening velocity of every muscle.
It holds the model's default of 10 optimal fiber lengths per second until $3\times10^{6}$ decision steps, then raises it in unit increments to 20 over the following $150\times10^{6}$ simulation steps.
Relieving the force--velocity limit this way pushes the top speed to about $6.5\,$m/s (Figure~\ref{fig:rsi}).
Maximum shortening velocity varies with fiber-type composition and is higher in muscles dominated by fast-twitch fibers, so we do not claim the doubled value is implausible. However, the maximum shortening velocity is not calibrated to any measured population.
We therefore treat this run as a diagnostic rather than as a performance result.

\end{document}